\pdfoutput=1

\documentclass[11pt]{article}

\usepackage[final]{acl}

\usepackage{times}
\usepackage{latexsym}

\usepackage[T1]{fontenc}

\usepackage[utf8]{inputenc}

\usepackage{microtype}

\usepackage{inconsolata}

\usepackage{graphicx}
\usepackage{amssymb}
\usepackage{booktabs}
\usepackage{amsthm, amscd, amsfonts, multirow, array}
\usepackage{makecell}
\usepackage{IEEEtrantools}
\usepackage{amsfonts}
\usepackage{amsmath}
\usepackage{enumitem}
\usepackage{colortbl}
\usepackage{textcomp}
\usepackage{makecell}
\usepackage{tabularx}
\usepackage[textsize=tiny]{todonotes}
\newlength\savewidth\newcommand\shline{\noalign{\global\savewidth\arrayrulewidth\global\arrayrulewidth 1pt}\hline\noalign{\global\arrayrulewidth\savewidth}}

\definecolor{baselinecolor}{gray}{.92}

\definecolor{demphcolor}{gray}{.2}

\definecolor{demphcolorinline}{gray}{.3}

\definecolor{demphcolor1}{gray}{.6}

\definecolor{eva01purple}{RGB}{168,119,200}

\definecolor{eva01green}{RGB}{82,208,83}

\definecolor{eva02red}{RGB}{236,35,35}

\definecolor{eva02yellow}{RGB}{249,157,83}

\definecolor{02pink}{RGB}{240,178,188}

\definecolor{Graylight}{gray}{0.9}
\definecolor{lightred}{RGB}{241,140,142}
\definecolor{00blue}{RGB}{100,149,237}

\newcommand{\ph}[1]{\textcolor{white}{#1}}

\newcommand*{\affaddr}[1]{#1} 
\newcommand*{\affmark}[1][*]{\textsuperscript{#1}}
\newcommand{\email}[1]{\href{mailto:#1}{\texttt{#1}}}

\title{Investigating and Mitigating Object Hallucinations in Pretrained Vision-Language (CLIP) Models}

\author{
Yufang Liu\textsuperscript{\rm 1}\thanks{~~Equal contribution.}, Tao Ji\textsuperscript{\rm 2,3}\footnotemark[1], Changzhi Sun\textsuperscript{\rm 1}, Yuanbin Wu\textsuperscript{\rm 1}, Aimin Zhou\textsuperscript{\rm 1}  \\[4pt]
\affaddr{\affmark[1]School of Computer Science and Technology, East China Normal University} \\
\affaddr{\affmark[2] School of Computer Science, Fudan University} \\
\affaddr{\affmark[3] Pazhou Laboratory, Huangpu} \\
\email{yfliu.antnlp@gmail.com}\quad\email{taoji@fudan.edu.cn}\quad\email{ybwu@cs.ecnu.edu.cn} \\
}

\begin{document}
\maketitle
\begin{abstract}
Large Vision-Language Models (LVLMs) have achieved impressive performance, yet research has pointed out a serious issue with object hallucinations within these models. 
However, there is no clear conclusion as to which part of the model these hallucinations originate from. 
In this paper, we present an in-depth investigation into the object hallucination problem specifically within the CLIP model, which serves as the backbone for many state-of-the-art vision-language systems. 
We unveil that even in isolation, the CLIP model is prone to object hallucinations, suggesting that the hallucination problem is not solely due to the interaction between vision and language modalities.
To address this, we propose a counterfactual data augmentation method by creating negative samples with a variety of hallucination issues.
We demonstrate that our method can effectively mitigate object hallucinations for the CLIP model, and we show that the enhanced model can be employed as a visual encoder, effectively alleviating the object hallucination issue in LVLMs.
\footnote{Our benchmark and code are publicly available on \url{https://github.com/Yufang-Liu/clip_hallucination}.}

\end{abstract}

\section{Introduction}
\label{sec:intro}


Current Large Vision-Language Models (LVLMs) demonstrate significant potential in tasks requiring joint visual and linguistic perception, 
such as image captioning \cite{agrawal2019nocaps}, 
visual question answering \cite{antol2015vqa}, 
visual grounding \cite{yu2016modeling}, 
and autonomous agents \cite{durante2024agent,xi2023rise}.
Despite the success of LVLMs, 
previous studies have revealed that they commonly suffer from hallucinations in practice, 
including object hallucinations \cite{DBLP:conf/emnlp/LiDZWZW23,DBLP:journals/corr/abs-2311-16922,DBLP:journals/corr/abs-2310-00754}, 
spatial hallucinations \cite{kamath2023s}, 
attribute hallucinations \cite{zhang2024debiasing}, etc.
It is widely believed that hallucinations degrade model performance and reliability, and severely impair the user experience in real-world applications
\cite{DBLP:journals/csur/JiLFYSXIBMF23}.

In this work, 
we focus on investigating the causes of the highly-concerned \textit{object hallucinations}, 
i.e.,  LVLMs generate nonexistent objects in the image \cite{9706727}.
A typical LVLM utilizes a Large Language Model (LLM) as its cognitive foundational model 
and employs a pre-trained image encoder as its visual perception module (mainly the CLIP encoder).
\citet{kamath2023s} investigated the spatial hallucination 
(e.g., confusing ``left of'' and ``right of'') in LVLMs, 
and they found that various CLIP encoders struggle to recognize simple spatial relationships 
(achieving only a 55.0\% accuracy on benchmarks, whereas humans are 98.8\%).
Inspired by their findings, we hypothesize that the CLIP visual encoder might also be one of the causes of object hallucinations.

Hence, we first curate the \textbf{O}bject \textbf{H}allucination \textbf{D}etection (\textbf{OHD}-Caps) benchmark from subsets of the COCO \cite{DBLP:conf/eccv/LinMBHPRDZ14}, Flickr30K \cite{DBLP:journals/tacl/YoungLHH14}, and Nocaps (as an out-of-domain benchmark because it comprises unseen objects) \cite{DBLP:conf/iccv/AgrawalAD0CJ0BP19} image caption datasets respectively, to more strictly measure the extent of object hallucinations present in CLIP encoders.
We randomly select 16k/1k/1.5k (train/dev/test) samples, with each sample containing one image, one positive descriptive text, and 27 negative descriptive texts.
The negative samples are perturbations of the positive sample, achieved by \textit{adding} descriptions of nonexistent objects or \textit{reducing} descriptions of existing objects.
Theoretically, a CLIP model without object hallucinations should accurately assign the highest CLIP score to the positive sample.
However, taking the most commonly used ``CLIP ViT-L/14'' in LVLMs as an example, it only scores the highest for positive samples in 19.0\% of cases.
Since we have observed that the CLIP encoder already has a serious issue with object hallucination, how can we mitigate it?

In the contrastive pretraining of CLIP, negative samples come from text descriptions of other images within the batch, which makes the distinction between them quite straightforward. 
However, mitigating object hallucinations requires the CLIP encoder to be able to differentiate between subtle errors at the object level.
We further fine-tune the CLIP model using the training set from \textbf{OHD}-Caps. 
By incorporating a fine-grained object-level contrastive loss, we greatly reduce object hallucinations in the CLIP. 
Then employing the fine-tuned CLIP as the visual encoder, the object hallucinations in our retrained LVLM, LLaVA-1.5, are also diminished.

In this paper, we study the object hallucinations of CLIP models.
Our main contributions are,
\begin{itemize}[leftmargin=*]
    \item we propose a benchmark, \textbf{OHD}-Caps, for evaluating object hallucinations in CLIP models. 
    \item we quantitatively evaluate a wide range of encoders from the CLIP family and find that they all exhibit severe object hallucination issues.
    \item we propose a fine-grained object-level contrastive loss to further fine-tune the CLIP model, significantly alleviating its object hallucination issues (e.g., from 14.3 to 82.5 for ``CLIP ViT-B/32'') and concurrently reducing the hallucination problems of the LLaVA-1.5 (from 80.2 to 83.2 on Nocaps), which uses it as a visual encoder.
\end{itemize}
\section{Related Work}

\subsection{Large Vision-Language Model}

Recently, inspired by the success of large language models (LLMs), researchers have begun to dedicate efforts to enhance vision language models (VLMs) by integrating robust LLMs, aiming to broaden the knowledge scope of the model and amplify its linguistic comprehension capabilities.

LVLM architectures typically consist of three components: a visual encoder, a modality connection module, and a LLM\@. 
The visual encoder and LLM are typically fixed large pretrained models, the visual encoder is usually a variant of the CLIP model~\cite{DBLP:conf/icml/RadfordKHRGASAM21}, used for extract visual features, while the LLM, such as LLaMA~\cite{DBLP:journals/corr/abs-2302-13971} and Vicuna~\cite{vicuna2023}, is used to integrate image information and text information, and completes the prediction of the target.
Research focuses on optimizing modality connection modules, with approaches like Flamingo's~\cite{DBLP:conf/nips/AlayracDLMBHLMM22} cross-attention module, LLaVA's~\cite{DBLP:conf/nips/LiuLWL23a} linear layer, and BLIP2's~\cite{DBLP:conf/icml/0008LSH23} Q-former, diverse yet all boosting VLM performance on various vision-language tasks.

\subsection{Hallucination in LVLMs}
Despite the fact that LVLMs perform well in solving visual-language tasks, they are also plagued by hallucinations. 
The problem of hallucinations in LVLMs mainly refers to the mismatch between visual input and textual output.
For example, in the image captioning task, hallucination refers to the generation of captions that describe objects that do not exist in the image.
Although the hallucination problem of LLMs has been widely studied in the NLP field~\cite{DBLP:journals/csur/JiLFYSXIBMF23}, there has not been enough research on mitigating the hallucination issue in LVLMs~\cite{DBLP:conf/acl/ShekharPKHNSB17, DBLP:journals/corr/abs-2402-00253, DBLP:journals/corr/abs-2310-14566}.
Recent efforts to mitigate hallucination in LVLMs have focused on enhancing each compoment of the model. 
For example,~\citet{DBLP:journals/corr/abs-2306-14565, DBLP:journals/corr/abs-2309-02301} constuct instruction-tuning datasets with contrastive question-answer pairs for LVLMs; 
~\citet{DBLP:journals/corr/abs-2309-14525, DBLP:journals/corr/abs-2312-00849} employ Reinforcement Learning from Human Feedback (RLHF) ~\cite{DBLP:conf/nips/StiennonO0ZLVRA20} to enchance the connection module between the modalities;
~\citet{DBLP:journals/corr/abs-2311-16922} propose a visual contrastive decoding strategy for LLM decoing.
Despite the wide application of the CLIP model in VLMs and its in-depth study in pairwise comparison context~\cite{DBLP:conf/iclr/Yuksekgonul0KJ023, DBLP:conf/nips/HsiehZMKK23}, there has been little discussion on its evaluation regarding hallucinations. Our research addresses this gap in the literature.
\section{The OHD-Caps Benchmark}\label{section:benchmarks}

\begin{figure*}[!htbp]
    \centering
    \includegraphics[width=0.95\textwidth]{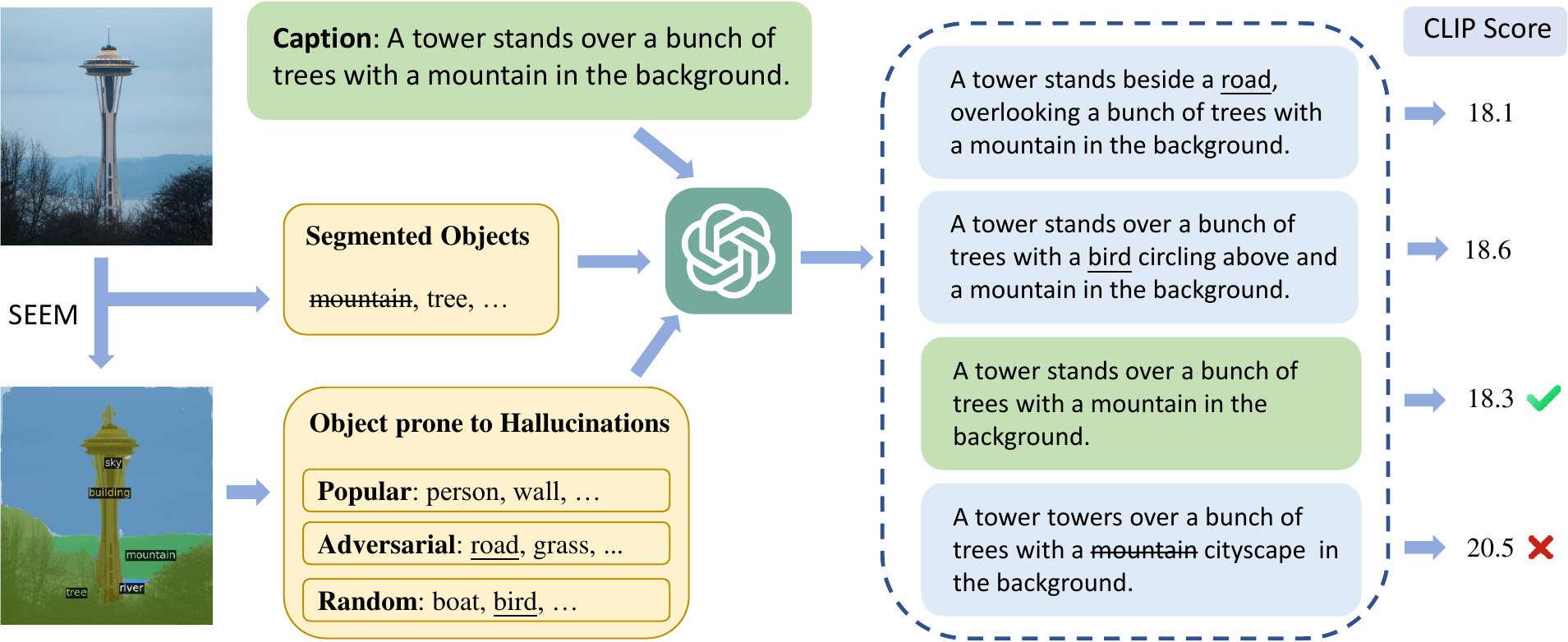}
    \caption{The pipeline of our benchmark creation process. For an image, we first use SEEM~\cite{DBLP:conf/nips/ZouYZLLWWGL23} to identify objects within the image and obtain illusory objects that do not exist in the picture through different sampling strategies. Then we ask GPT to insert or delete objects in the original sentences to create negative samples. We provide both positive and negative samples to the CLIP model to observe if the model predicts the positive samples as having the highest score. This image is from the NoCaps dataset, and the model is CLIP ViT-B/32.}
    \label{fig:dataset}
\end{figure*}

Recent studies have found that LVLMs are prone to object hallucinations~\cite{DBLP:conf/emnlp/LiDZWZW23, DBLP:journals/corr/abs-2310-00754}. In response, researchers have developed several datasets to assess the extent of these hallucinations in such models~\cite{DBLP:conf/emnlp/LiDZWZW23, DBLP:journals/corr/abs-2308-15126}. However, there is a relative lack of assessment work regarding the hallucinatory effects of the CLIP model, which is widely used as a visual encoder within LVLMs.
In this section, we introduce the \textbf{O}bject \textbf{H}allucination \textbf{D}etection benchmark (OHD-Caps) we create to evaluate the object hallucination problem in CLIP models and the pipeline for evaluations. 
Figure~\ref{fig:dataset} shows the pipeline of our benchmark creation process.

\subsection{Dataset Construction}

CLIP is a versatile neural network that excels at image understanding and can predict text for images in a zero-shot manner.
To evaluate the CLIP model's ability to handle object hallucinations in paired comparison scenarios, given an image with a correct caption, we create incorrect captions containing hallucinatory content. The purpose is to observe whether the model can accurately select the correct text without hallucinations.

\paragraph{Inserting Hallucinatory Objects}
Previous work~\cite{DBLP:conf/emnlp/LiDZWZW23, DBLP:journals/corr/abs-2310-00754} show that LVLMs are more prone to generate hallucinatory responses for objects that frequently appear in the dataset.
Inspired by this, we create negative samples by inserting objects prone to hallucination into the correct captions.
To collect object annotations, we first use SEEM~\cite{DBLP:conf/nips/ZouYZLLWWGL23} to automatically segment objects in the images. 
Three kinds of hallucinatory objects are collected: \textit{random objects} which are sampled randomly, \textit{popular objects} which are the top frequent objects in the whole dataset, and \textit{adversarial objects} which are the top frequent objects with the segmented objects.
Each category contains three objects. To create examples with varying levels of hallucinations, we attempt to insert one to three objects for each category, resulting in each type of hallucination containing a total of 7 ($\sum_{r=1}^3 C_3^r$) samples.

Given a caption text and several hallucinatory objects, we insert the objects into the appropriate locations in the caption, which can be effectively achieved with the help of GPT4.
Automatically, the caption and objects are fed to the GPT4, with the prompt as \textit{Add\_Prompt} (see Table \ref{tab:gpt4_prompt}).

\paragraph{Removing existing Objects}
Except from inserting hallucinatory objects, we also remove objects from the captions to create negative samples.
We randomly select 1 or 2 segmented objects in the image which results in 6 negative samples ($\sum_{r=1}^2 C_3^r$), and ask GPT4 to remove them from the caption with the \textit{Remove\_Object\_Prompt}. 
To account for scenarios where the identified objects are not present in the title text, we ask GPT to alter elements like objects, colors, and properties in the original caption, the prompt we use is \textit{Alter\_Object\_Prompt}.
The prompt can be found in Table \ref{tab:gpt4_prompt}.

we construct a dataset of 500 samples for each of the COCO~\cite{DBLP:conf/eccv/LinMBHPRDZ14}, Flickr30K~\cite{DBLP:journals/tacl/YoungLHH14}, and the out of domain subset of NoCaps Validation datasets~\cite{DBLP:conf/iccv/AgrawalAD0CJ0BP19}, with 27 negative samples for each image. 
Specifically, the out of domain subset of NoCaps comprises objects not seen in the COCO dataset, commonly used to measure a model's ability to generalize to unseen classes. \footnote{Our selection of Nocaps as the out-of-domain dataset is specific to our fine-tuning process in Section \ref{sec:methology} and not the pre-training process of CLIP.}
The average length of the captions in the datasets is shown in Table~\ref{tab:dataset}.

\subsection{Evaluation and Analysis}

We study several models to evaluate their performance on our benchmark.
Each image is paired with a correct caption and 27 negative samples, 
and models are required to calculate the similarity between the image and the caption candidates and select the correct caption.

\paragraph{Models}
We evaluate a variety of models on our benchmark, including  CLIP~\cite{DBLP:conf/icml/RadfordKHRGASAM21} ViT-B/32 and ViT-L/14; 
MetaCLIP~\cite{DBLP:journals/corr/abs-2309-16671} and DFN2B CLIP~\cite{DBLP:journals/corr/abs-2309-17425} are models pretrained on high-quality dataset after data curation;
CLIPA\cite{DBLP:journals/corr/abs-2306-15658} which achieves efficient training by using shorter image/text sequences, which reduces the computational load during the training period;
EVA CLIP~\cite{DBLP:journals/corr/abs-2303-15389} which employs innovative representation learning technology, optimization methods, and enhancement strategies to improve model performance;
SigLIP\cite{DBLP:conf/iccv/ZhaiM0B23} which employs a contrastive learning loss function based on the Sigmoid function instead of the traditional softmax for pre-training on language and image data;
CLIP ConvNext\cite{DBLP:conf/cvpr/0003MWFDX22} is a variant of the CLIP model that uses ConvNext as the image encoder;
CLIP NLLB-SigLip~\cite{DBLP:journals/corr/abs-2309-01859} is another variant that combines a text encoder from the NLLB model~\cite{DBLP:journals/corr/abs-2207-04672} and an image encoder from the SigLIP model;
NegCLIP~\cite{DBLP:conf/iclr/Yuksekgonul0KJ023}, an improved model based on CLIP ViT-B/32, which enhances the understanding of relationships between objects, attributes, and the sequence of words by swapping phrases;
CECLIP~\cite{DBLP:journals/corr/abs-2306-08832} which further develop enhanced negative samples and employ contrastive loss to enhance compositional reasoning;
FLAVA~\cite{DBLP:conf/cvpr/SinghHGCGRK22} which is a single unified foundation model which can work across vision, language as well as vision-and-language multi-modal tasks;
CoCa~\cite{DBLP:journals/tmlr/YuWVYSW22} is a pretrained model with contrastive and generative learning objectives;
XVLM~\cite{DBLP:journals/corr/abs-2111-08276} which aligns the visual concept and textual input in a multi-grained manner with 14M and 16M pretrained images;
BLIP~\cite{DBLP:conf/icml/0001LXH22} which effectively utilizes the noisy web data by bootstrapping the captions with 14M and 129M pretrained images;
BLIP2~\cite{DBLP:conf/icml/0008LSH23}~\footnote{We use the image-text matching head for both BLIP and BLIP2.} which bridges the gap between the visual and textual modalities with a Q-former.

\paragraph{Results}
\begin{table}[t]
    \centering
    \setlength{\tabcolsep}{3pt}
    \resizebox{\columnwidth}{!}{  
    \begin{tabular}{lrcccc}
    \toprule
        \multirow{2}{*}{\bf Model} & \multirow{2}{*}{\bf Params}&\multicolumn{4}{c}{\bf OHD-Caps Benchmark} \\
        \cmidrule{3-6}
        &&\bf COCO & \bf Flickr30K & \bf NoCaps &\bf Avg.\\
        \midrule
        \multicolumn{6}{c}{ (a) comparisons with CLIP Models} \\
        \midrule
        CLIP ViT-B/16      & 149M & 16.6 & 17.2 & 8.6  & 14.1  \\
        CLIP ViT-B/32      & 151M & 15.2 & 17.6 & 10.2 & 14.3  \\
        CLIP ViT-L/14      & 428M & 22.4 & 22.6 & 12.0 & 19.0  \\
        MetaCLIP B/32      & 151M & 25.6 & 25.2 & 16.0 & 22.3  \\
        MetaCLIP L/14      & 428M & 36.8 & 26.4 & 19.4 & 27.5  \\
        CLIPA V2 L/16      & 428M & 35.6 & 31.0 & 18.8 & 28.5  \\
        EVA-02 CLIP-B/16   & 149M & 26.4 & 25.4 & 18.6 & 23.5  \\
        EVA-02 CLIP-L/14   & 428M & 38.8 & 31.6 & 21.4 & 30.6  \\
        DFN2B CLIP B/16    & 149M & 29.4 & 27.8 & 17.0 & 24.7  \\
        DFN2B CLIP L/14    & 428M & 37.6 & 37.8 & 23.2 & 32.9  \\
        CLIP ConvNext-B    & 180M & 34.0 & 28.0 & 20.4 & 27.5  \\
        CLIP ConvNext-L    & 352M & 43.4 & 35.8 & 25.0 & 34.7   \\
        SigLIP B/16        & 203M & 34.2 & 32.2 & 23.8 & 30.1  \\
        SigLIP L/16        & 652M & 48.4 & 38.4 & \bf 30.8 & 39.2   \\
        SigLIP SoViT-400m  & 877M & 50.8 & \bf 41.4 & 26.6 & \bf 39.6  \\
        CLIP NLLB-SigLip-B & 508M & 25.2 & 20.0 & 22.6 & 22.6  \\
        CLIP NLLB-SigLip-L & 1.1B & 32.6 & 29.0 & 26.4 & 29.3  \\
        NegCLIP            & 151M & 32.8 & 28.0 & 25.0 & 28.6 \\
        CECLIP             & 151M & \bf 52.8 & 40.8 & 23.4 & 39.0 \\
        \midrule
        \multicolumn{6}{c}{ (b) comparisons with other Image-Text Matching Models} \\
        \midrule
        FLAVA              & 350M & 28.0 & 28.4 & 16.6 & 24.3 \\
        CoCa               & 2.1B & 26.0 & 24.4 & 20.0 & 23.5 \\
        XVLM 4M            & 216M & 46.4 & 35.8 & 34.0 & 38.7 \\
        XVLM 16M           & 216M & 41.8 & 19.4 & 21.8 & 27.7 \\
        BLIP 14M           & 583M & 51.4 & \bf 48.0 & \bf 42.0 & 47.1 \\
        BLIP 129M          & 583M & 40.8 & 38.0 & 31.2 & 36.7 \\
        BLIP2              & 3.4B & \bf 62.6 & 42.2 & 41.2 & \bf 48.7 \\
        \bottomrule
    \end{tabular}
    }
    \caption{Results of various models on our benchmark. NoCaps subset is used to evaluate zero-shot generalization.
    }
    \label{tab:main_results}
\end{table}

Table~\ref{tab:main_results} shows the results of the models on our benchmark.
From the results, we could find that,
\begin{itemize}[leftmargin=*]
    \item First of all, the vanilla CLIP models perform poorly across all three datasets, indicating their limited ability to recognize illusory objects in images. 
    Multiple variants of CLIP, through improvements in data (e.g., MetaCLIP, DFN2B CLIP), model architecture (e.g., CLIP ConvNext, CLIP NLLB-SigLip), and training methods (e.g., CLIPA, EVA CLIP, SigLip), achieve a slight enhancement in the performance of the original CLIP. Among these variants, SigLIP demonstrates the most notable performance, exhibiting the best results on out-of-domain datasets and showcasing superior generalization capabilities.
    
    \item Secondly, NegCLIP attempts to enhance the model's understanding of the text by parsing and substituting phrases, but it only achieves a marginal improvement compared to the original CLIP model. 
    CECLIP exhibits relatively better performance, which is mainly due to the constructed negative samples enhancing the model's comprehension of the combined semantics of sentences.
    The NegCLIP and CECLIP models are trained on the COCO training set to distinguish between positive samples and enhanced negative samples. This might contribute to CECLIP's good performance on the COCO dataset, owing in part to the model's memory of the original correct text. 
    However, their performance on the NoCaps dataset indicates that these models cannot effectively differentiate hallucinated objects.
    
    \item Furthermore, generative vision-language models typically achieve higher performance than vanilla CLIP models due to their more precise alignment of image and text representations. Furthermore, it is generally observed that the larger the model parameters, the better the performance. In particular, BLIP2, which has the highest number of parameters, performs best across all three datasets. In comparison, the XVLM 4M model has relatively fewer parameters but still demonstrates good performance. This indicates that XVLM's strategy of multi-scale alignment indeed assists the model in more accurately capturing the fine-grained details within images.
    
    \item Finally, the overall trend among different models is consistent across the three datasets, with their performance typically being the lowest on the NoCaps dataset. Although fewer objects are recognized on the NoCaps dataset than Flickr30K, the performance is the lowest there due to the inclusion of categories that are out-of-domain. The BLIP 14M model demonstrates the best performance on both Flickr and NoCaps, which indicates its strong generalization capabilities.
\end{itemize}

\paragraph{Analysis}
The inability of models to recognize hallucinated objects primarily stems from the data used and the learning methods employed. 
The vanilla CLIP model is trained with a large number of image-caption pairs collected from the internet, using a contrastive loss function for optimization. 
Those captions are often brief and noisy, and the model is optimized to differentiate between correct and a multitude of incorrect image-text pairs. 
However, because the incorrect pairs are usually significantly different from the correct ones, the model can easily distinguish them. This means that the model does not need to learn the rich details in the pictures to make accurate predictions.
To address this issue, we need to make improvements to the original CLIP model in terms of data utilization and learning methodologies.

\section{Methodology} \label{sec:methology}

We first revisit the training process of the vanilla CLIP model.
Let $I$ be the image and $T$ be the text, 
the training objective of CLIP is to maximize the similarity between the image and text pairs and minimize the similarity between the image and text pairs that are not matched. 
The loss function is defined as:
\begin{equation}
    \begin{aligned}
        \mathcal{L}_{i2t} &=  -\log \frac{\exp(I \cdot T^+ / \tau)}{\sum_{T^* \in \{T^+, T^-\}} \exp(I \cdot T^* / \tau)}, \\
        \mathcal{L}_{t2i} &=  -\log \frac{\exp(T \cdot I^+ / \tau)}{\sum_{I^* \in \{I^+, I^-\}} \exp(T \cdot I^* / \tau)}, \\
        \mathcal{L}_0 &= \frac{1}{2} \big( \mathcal{L}_{i2t} + \mathcal{L}_{t2i} \big),
    \end{aligned}
\end{equation}
where $T^+$ and $I^+$ are the correct text and image, and $T^-$ and $I^-$ are the incorrect text and image, respectively.

With the addition of the negative samples $T^{neg}$ created as in the previous section, we could modify the loss $\mathcal{L}_{i2t}$ as:
\begin{equation}
    \mathcal{L}_{i2t} = -\log \frac{\exp(I \cdot T^+ / \tau)}{\sum_{T^* \in \{T^-, T^{neg}, T^+\}} \exp(I \cdot T^* / \tau)}.
\end{equation} 

To further enhance the model's ability to distinguish between positive and negative samples, we additionally introduce a margin loss. 
This is to ensure that the distance between an image and its corresponding correct text is smaller than the distance to incorrect text by a specific threshold. 
This concept can be formulated as:
\begin{equation}
    \mathcal{L}_1 = \max(0, \tau_1 - I \cdot T^+ + I \cdot T^*), 
\end{equation}
where $\tau_1$ is the margin threshold, and $T^*=\{T^-, T^{neg}\}$.

Additionally, we generate enhanced negative samples by introducing perturbations to the original positive samples. Such negative samples are typically more challenging to distinguish than other negative samples within the batch. To encourage the model to recognize the partially correct information contained in the enhanced negative samples, resulting in a higher similarity to the positive samples compared to other negative samples within the batch, we introduce a margin loss between the in-batch negative samples and the enhanced negative samples:
\begin{equation}
    \mathcal{L}_2 = \max(0, \tau_2 - I \cdot T^{neg} + I \cdot T^{-}),
\end{equation}
where $\tau_2$ is the margin threshold. 

Next, we assign different weights to the aforementioned loss terms, allowing the model to learn adaptively. Consequently, the final loss function can be expressed as follows:
\begin{equation}
    \mathcal{L} = \frac{1}{2} \big( \mathcal{L}_{t2i} + \mathcal{L}_{i2t} \big )+ \lambda_1 \mathcal{L}_1 + \lambda_2 \mathcal{L}_2. 
\end{equation}

\begin{table}[t]
    \centering
    \resizebox{0.9\linewidth}{!}{  
    \begin{tabular}{lcccc}
    \toprule
    \multirow{2}{*}{\bf Model} &\multicolumn{4}{c}{\bf OHD-Caps } \\
    \cmidrule(lr){2-5}
    & \bf COCO &\bf Flickr30k   &\bf NoCaps &\bf Avg.\\

    \midrule
    Random & 3.6 & 3.6 & 3.6 & 3.6 \\
    \midrule
    \multicolumn{5}{c}{ (a) comparisons with CLIP-\textbf{Base} baselines} \\
    \midrule
    CLIP-B/32   & 15.2 & 17.6 & 10.2 & 14.3 \\
    NegCLIP         & 32.8 & 28.0 & 25.0 & 28.6 \\
    CECLIP          & 52.8 & 40.8 & 23.4 & 39.0 \\
    \textbf{Ours}-B/32            & \bf 80.4  &\bf 85.0 & \bf 82.0  & \bf 82.5 \\
    \midrule
    \multicolumn{5}{c}{ (b) comparisons with CLIP-\textbf{Large} baselines} \\
    \midrule
    CLIP-L/14 &26.0 & 27.0&16.8&23.3 \\
    \textbf{Ours}-L/14            & \bf87.0&\bf91.0 &\bf 88.4&\bf 88.8  \\
    \bottomrule
    \end{tabular}
    }
    \caption{Results on OHD-Caps. CLIP-B/32, and CLIP-L/14 represent CLIP ViT-B/32 and CLIP ViT-L/14 336 px respectively.}
        \label{tab:main_finetune}
\end{table}

\section{Experiments}
\paragraph{Training Datasets}
We sample 8k images from the training set of COCO and 8k images from Flickr30k datasets, then generate negative samples for each image as in Section~\ref{section:benchmarks}.
Additionally, we randomly select $\sim$1k samples from the COCO dataset's validation set as our dev set for the selection of hyper-parameters.
Detailed information about the dataset is provided in Table \ref{tab:dataset}.

\paragraph{Training Details}
We utilize the CLIP ViT/32-B and CLIP ViT/14-L-336px implemented by Huggingface~\cite{wolf-etal-2020-transformers} as the initial models and conduct fine-tuning for 10 epochs. 
The training process is carried out on a single A6000 GPU, with batch sizes of 56 and 14 set for the base and large models, respectively, and the learning rate is set at 1e-6.
The selection of hyper-parameters is determined by their performance on the validation set, where $\lambda_1$ and $\lambda_2$ are set as 0.1 and 0.1, $\tau_1$ and $\tau_2$ are set as 2.

\paragraph{Evaluation}
To verify the impact of our method on the model's generalization capabilities, we conducted zero-shot experiments on the following datasets: CIFAR-10/100~\cite{cifar}, ImageNet-1K ~\cite{deng2009imagenet}, DTD~\cite{cimpoi14describing}, Eurosat~\cite{helber2019eurosat}, GTSRB~\cite{stallkamp2012man} and STL10~\cite{coates2011analysis}.

\begin{table}[t]
\centering
\resizebox{\linewidth}{!}{ 
     \begin{tabular}{l|cccccc|c}
        &
        \rotatebox[origin=l]{90}{CIFAR-10~(\citeyear{cifar})} &
        \rotatebox[origin=l]{90}{CIFAR-100~(\citeyear{cifar})} & 
        \rotatebox[origin=l]{90}{ImageNet~(\citeyear{deng2009imagenet})} &
        \rotatebox[origin=l]{90}{Eurosat~(\citeyear{helber2019eurosat})} & 
        \rotatebox[origin=l]{90}{GTSRB~(\citeyear{stallkamp2012man})} & 
        \rotatebox[origin=l]{90}{STL10~(\citeyear{coates2011analysis})} & 
        \rotatebox[origin=l]{90}{\ph{.}\textbf{avg. top-1 acc.}}
        \\
        \shline

        \multicolumn{8}{c}{ (a) comparisons with CLIP-\textbf{Base} baselines} \\
        \hline
         CLIP-B/32 & \bf 89.8  & 64.2  & \bf 63.3 & 46.3  & \bf 32.6  & \bf 97.1  &  65.6  \\
         NegCLIP   & 85.9  & 60.9  & 55.7 & 31.9  & 26.8  & 95.8  &  55.8  \\
         CECLIP    & 81.1  & 55.0  & 40.4 & 41.9  & 20.6  & 95.6  &  59.5  \\
         \textbf{Ours}-B/32      & 89.1  & \bf 66.0  & 60.5  & \bf 51.7  & 31.9  & 96.5  & \bf 66.0   \\
        \shline
        
        \multicolumn{8}{c}{ (b) comparisons with CLIP-\textbf{Large} baselines} \\
        \hline
         CLIP-L/14     & 95.0  & 74.4  & \bf 76.6   & 61.4  & \bf52.4  & 99.4  &  \bf 76.5  \\
         \textbf{Ours}-L/14     & 95.0  & \bf 74.8  & 72.8   & \bf 67.3  & 43.6  & 99.4  & 75.5   \\
        \shline
        \end{tabular}
        }
        \caption{Zero-shot results on various datasets. The last column displays the average performance across 7 datasets.}
        \label{tab:zero_shot}
\end{table}
\subsection{Main Results}
We present the results for our self-constructed dataset in Table \ref{tab:main_finetune}, and various zero-shot datasets in Table \ref{tab:zero_shot}.
From the results, we could find:
\begin{itemize}[leftmargin=*]
    \item Our model shows comparable zero-shot performance to vanilla CLIP Models (65.6 vs 66.0) and achieves significant improvements in hallucination recognition (14.3 vs 82.5). 
    NegCLIP and CECLIP enhance the model's capability of understanding composites by constructing negative samples and also achieve a moderate improvement on the OHD-Caps benchmark, with performance rising from 14.3\% to 39.0\%.
    However, the zero-shot performance of NegCLIP and CECLIP significantly decreases. This could be due to their reliance on rule-based methods to construct negative samples (such as swapping phrases), which may interfere with the model's understanding of sentence semantics.
    
    \item Our model also demonstrates strong generalization capabilities in hallucination recognition. NegCLIP, CECLIP, and our model are all fine-tuned on the training set of the COCO dataset. 
    Although they show varying degrees of performance improvement in COCO-related hallucination tests (NegCLIP at 32.8\%, CECLIP at 52.8\%), their performances are worse when facing unknown categories (NegCLIP at 25.0\%, CECLIP at 23.4\% for NoCaps images), indicating limited generalization capabilities of the models. 
    In contrast, our model performs consistently across three different datasets, at approximately 82\%. 
    This result verifies that our model can effectively distinguish hallucinated objects in different datasets and possesses the capability to generalize across datasets.
    
\end{itemize}


\begin{table}[t] 
    \centering
    \setlength{\tabcolsep}{4pt}
    \resizebox{\columnwidth}{!}{ 
    \begin{tabular}{llcccc}
    \toprule 
    \multirow{2}{*}{\bf Dataset} & \multirow{2}{*}{\bf Criterion} & \multicolumn{2}{c}{\bf Full Fine FT } & \multicolumn{2}{c}{\bf LoRA FT } \\
    \cmidrule(lr){3-4} \cmidrule(lr){5-6}
    && \bf LLaVA & \bf Ours & \bf LLaVA & \bf Ours \\
    \midrule
    \multirow{5}{*}{COCO}
    & Accuracy ($\uparrow$)   &\bf 85.4&81.2&85.7&\bf88.3  \\
    & Precision ($\uparrow$)  &81.8&\bf 90.9&81.8&\bf89.7  \\
    & Recall ($\uparrow$)     &\bf 91.9&85.1&\bf92.5&86.9  \\
    & F1 Score ($\uparrow$)   &86.4&\bf 87.9&86.7&\bf88.2  \\
    & Yes ($\rightarrow$50\%) &56.5&\bf 46.9&56.8&\bf48.6  \\
    \midrule
    \multirow{5}{*}{Flickr30K}
    & Accuracy ($\uparrow$)   &73.7&\bf81.2&74.4&\bf82.8  \\
    & Precision ($\uparrow$)  &67.5&\bf78.5&67.9&\bf83.0  \\
    & Recall ($\uparrow$)     &\bf96.9&88.0&\bf96.9&85.7  \\
    & F1 Score ($\uparrow$)   &79.2&\bf82.7&79.5&\bf83.5  \\
    & Yes ($\rightarrow$50\%) &73.1&\bf56.8&72.5&\bf52.9  \\
    \midrule
    \multirow{5}{*}{NoCaps}
    & Accuracy ($\uparrow$)   &76.7&\bf 81.3&76.7&\bf82.6  \\
    & Precision ($\uparrow$)  &71.2&\bf 80.6&71.2&\bf81.8  \\
    & Recall ($\uparrow$)     &\bf 92.7&84.0&\bf92.3&84.9  \\
    & F1 Score ($\uparrow$)   &80.2&\bf 82.0&80.2&\bf83.2  \\
    & Yes ($\rightarrow$50\%) &66.0&\bf 52.7&65.6&\bf52.3  \\
    \bottomrule
    \end{tabular}
    }
    \caption{Results on expanded POPE datasets. Yes denotes the proportion of answering ``Yes'' to the given question. 
    }
    \label{tab:POPE}
\end{table}

\begin{table}[t]
    \centering
    \setlength{\tabcolsep}{3pt}
    \resizebox{\linewidth}{!}{  
    \begin{tabular}{lcccccccc}
    \toprule
    \multirow{2}{*}{\bf Model} &\multicolumn{4}{c}{\bf Full FT }&\multicolumn{4}{c}{\bf LoRA FT } \\
    \cmidrule(lr){2-5} \cmidrule(lr){6-9}
    & \bf $C_S \downarrow$ &\bf $C_I \downarrow$  &Cover$\uparrow$ &Length&\bf $C_S \downarrow$ &\bf $C_I \downarrow$&Cover$\uparrow$&Length\\
    \midrule
    LLaVA   &56.4 & 14.9&79.1&106.4&58.2 &16.4&\bf79.9&106.5  \\
    \textbf{Ours}       &\bf55.0 &\bf14.5&\bf79.2&107.5 &\bf56.8 &\bf14.9&79.2&108.5 \\
    \bottomrule
    \end{tabular}
    }
    \caption{CHAIR hallucination evaluation results (max new tokens is 512) on COCO dev set. Smaller values correspond to less hallucinations.}
    \label{tab:chair}
\end{table}

\begin{table}[t] 
    \centering
    \setlength{\tabcolsep}{2pt}
    \resizebox{\columnwidth}{!}{ 
    \begin{tabular}{llcccc}
    \toprule 
    \multirow{2}{*}{\bf Dataset} & \multirow{2}{*}{\bf Criterion} & \multicolumn{2}{c}{\bf Full FT } & \multicolumn{2}{c}{\bf LoRA FT } \\
    \cmidrule(lr){3-4} \cmidrule(lr){5-6}
    && \bf LLaVA & \bf Ours & \bf LLaVA & \bf Ours \\
    \midrule
    \multirow{3}{*}{Generative}
    & $C_S$ ($\downarrow$)   &7.2&\bf6.5&7.2&\bf6.1  \\
    & $C_I$ ($\downarrow$)     &35.4&\bf31.7&33.4&\bf30.1  \\
    & Cover ($\uparrow$)  &\bf52.2&50.9&\bf51.7&50.7  \\
    \midrule
    \multirow{4}{*}{Discriminative}
    & Accuracy ($\uparrow$)   &74.3&\bf80.2&74.2&\bf80.8  \\
    & Precision ($\uparrow$)  &\bf93.9&85.5&\bf93.5&86.4  \\
    & Recall ($\uparrow$)     &65.6&\bf84.4&65.7&\bf84.3  \\
    & F1 ($\uparrow$)     &77.2&\bf84.9&77.2&\bf85.3  \\
    \bottomrule
    \end{tabular}
    }
    \caption{Results on AMBER dataset which includes the assessment of hallucinations in both discriminative and generative responses.}
    \label{tab:Amber}
\end{table}

\begin{table}[t] 
    \centering
    \setlength{\tabcolsep}{3pt}
    \resizebox{\columnwidth}{!}{ 
    \begin{tabular}{lcccccc}
    \toprule 
    \bf Model & \bf Existence & \bf Attribute	 & \bf State & \bf Number & \bf Action	&\bf Relation \\
    \midrule
    \multicolumn{7}{c}{(a) Full FT} \\
    \hline
     LLaVA &83.5&72.4&67.0&78.7&85.2&57.4 \\
    \bf Ours  &\bf 94.2&\bf 79.1&\bf77.1&\bf79.5&\bf88.6&\bf64.3 \\
    \midrule
    \multicolumn{7}{c}{(b) LoRA FT} \\
    \hline
    LLaVA &83.0&73.2&71.7&73.2&81.8&56.5  \\
    \bf Ours  &\bf94.3&\bf79.4&\bf77.8&\bf80.4&\bf86.7&\bf63.4 \\
    \bottomrule
    \end{tabular}
    }
    \caption{Detailed performance on AMBER discriminative subset which includes evaluation results of other types of hallucinations, such as attribute, number, and relation.}
    \label{tab:AMBER_supplemnt}
\end{table}

\begin{table}[t] 
    \centering
    \setlength{\tabcolsep}{3pt}
    \resizebox{\columnwidth}{!}{ 
    \begin{tabular}{lccccc}
    \toprule 
    \bf Model & \bf MME & \bf VQAv2 & \bf VisWiz & \bf SciQA-IMG & \bf TextVQA \\
    \midrule
    \multicolumn{6}{c}{(a) Full FT} \\
    \hline
     LLaVA &1459.4&79.1&48.9&\bf69.4&\bf58.5 \\
    \bf Ours  &\bf1487.2&\bf79.2&\bf50.0&69.3&58.2 \\
    \midrule
    \multicolumn{6}{c}{(b) LoRA FT} \\
    \hline
    LLaVA &1445.4&79.1&46.8&\bf69.8&\bf58.5  \\
    \bf Ours  &\bf1455.4&\bf79.2&\bf47.2&68&58.4 \\
    \bottomrule
    \end{tabular}
    }
    \caption{Results on various benchmarks.}
    \label{tab:other_bnechmark}
\end{table}

\subsection{Evaluation for LVLM}
To verify the effectiveness of the enhanced CLIP model compared to the original CLIP in assisting large vision-language models to mitigate the issue of object hallucination, we replace the CLIP ViT-L/14-336px baseline model in LLaVA-1.5 with our fine-tuned version.
We train LLaVA~\cite{DBLP:conf/nips/LiuLWL23a} from scratch using the hyper-parameters specified in the original paper. Comparison results with other methods, such as constructing SFT data~\cite{DBLP:journals/corr/abs-2311-07574} or introducing DPO processes~\cite{DBLP:journals/corr/abs-2402-11411, DBLP:journals/corr/abs-2311-16839} for further alignment can be found in Appendix \ref{sec:compare_sec}. 

\paragraph{Hallucination Detection}
To evaluate the occurrence of hallucination phenomena in discriminative and generative responses within models, we select the following evaluation methods for analysis: an extended version of the POPE dataset \cite{DBLP:conf/emnlp/LiDZWZW23} for discriminative response evaluation, and CHAIR evaluation \cite{DBLP:conf/emnlp/RohrbachHBDS18} for generative response; the AMBER dataset \cite{DBLP:journals/corr/abs-2311-07397} contains both types of evaluations. 
The format of the question contained in POPE is: `Is there a X in the image?', where X refers to the name of the object. 
The questions in the dataset are designed such that the objects are present and absent in equal measure, therefore the ideal `yes' response rate should be around 50\%. 
We extend the POPE dataset and incorporate the Flickr30k and NoCaps domains to test the model's generalization capabilities.
The CHAIR metric evaluates object hallucinations in image descriptions by measuring the ratio of referenced objects not found in the ground-truth label set, with CHAIR$_S$ for sentence level: 
$$
    C_S  =\frac{\mid\{\text { hallucinated objects }\} \mid}{\mid\{\text { all mentioned objects }\} \mid},
$$
CHAIR$_I$ for image-level analysis:
$$
    C_I =\frac{\mid\{\text { captions w/ hallucinated objects }\} \mid}{\mid\{\text { all captions }\} \mid},
$$
and Cover measures the object coverage of responses:
$$
   \text{Cover} = \frac{\mid\{\text { captions w/ hallucinated objects }\} \mid}{\mid\{\text { ground truth objects }\} \mid}.
$$

Table \ref{tab:POPE}, \ref{tab:chair}, \ref{tab:Amber} show the results of the expanded POPE dataset, CHAIR evaluation, and AMBER dataset, respectively.
From the results, we could find:
\begin{itemize}[leftmargin=*]
\item For discriminative responses, our model achieves significant improvements on various datasets. On the POPE dataset, compared to the original, it attains a better balance between accuracy and recall which results in a higher F1 score and also approaches a more ideal balance in the proportion of "Yes" responses. The same phenomenon of performance improvement is also observed in the AMBER dataset.
\item For generative responses, our model demonstrates a lower proportion of hallucinated content on the COCO validation set and the AMBER dataset, while maintaining a relatively stable coverage and response length.

\end{itemize}



\paragraph{General Performance}
We evaluate the model's general performance on different datasets, which include:
MME-Perception \cite{DBLP:journals/corr/abs-2306-13394} evaluates the model's visual perception with yes/no questions.
VQA-v2 \cite{DBLP:conf/cvpr/GoyalKSBP17} evaluate model's visual perception capabilities on open-ended short answers;
VizWiz \cite{DBLP:conf/cvpr/Gurari0SGLGLB18} and ScienceQA \cite{DBLP:conf/nips/LuMX0CZTCK22} with multiple choice to evaluate the model's zero-shot generalization on visual questions;
TextVQA \cite{DBLP:conf/cvpr/SinghNSJCBPR19} contains text-rich visual question answering.

Results are shown in Table \ref{tab:other_bnechmark}. 
We can observe that with full fine-tuning, there is a slight improvement in the model's average performance. Specifically, the average performance of the model across five datasets increased from 343.1 to 348.5, with the most notable improvement on the MME dataset. Conversely, when employing LoRA fine-tuning, the average performance of the model remained unchanged  (340.0 vs 341.7).

\subsection{Ablation Study}

\begin{table}
    \centering
    \setlength{\tabcolsep}{4pt}
    \resizebox{\linewidth}{!}{
        \begin{tabular}{lccccccc} \toprule
            \textbf{Model}  & $\mathcal{L}_0$  & $\mathcal{L}_1$&$\mathcal{L}_2$& \textbf{OHD-Caps}& \textbf{CIFAR10}&\textbf{CIFAR100} & \textbf{Avg.} \\ \midrule
            CLIP    &&&                          &14.3&89.8&64.2&39.4\\
            Ours& \checkmark &&                  &80.1&88.6&\bf66.4&79.1\\
            &  \checkmark&\checkmark&            &80.5&\bf89.3&66.0&79.4\\
            &  \checkmark&&\checkmark            &81.6&89.0&66.3&80.0\\
            &  \checkmark&\checkmark&\checkmark  &\bf 82.5&89.1&66.0&\bf80.5\\
            \bottomrule
        \end{tabular}}
        \caption{Ablation of losses on CLIP ViT-B/32. }
        \label{tab:ablation_loss}
\end{table}


\begin{figure}[!htbp]
    \centering
    \includegraphics[width=0.9\linewidth]{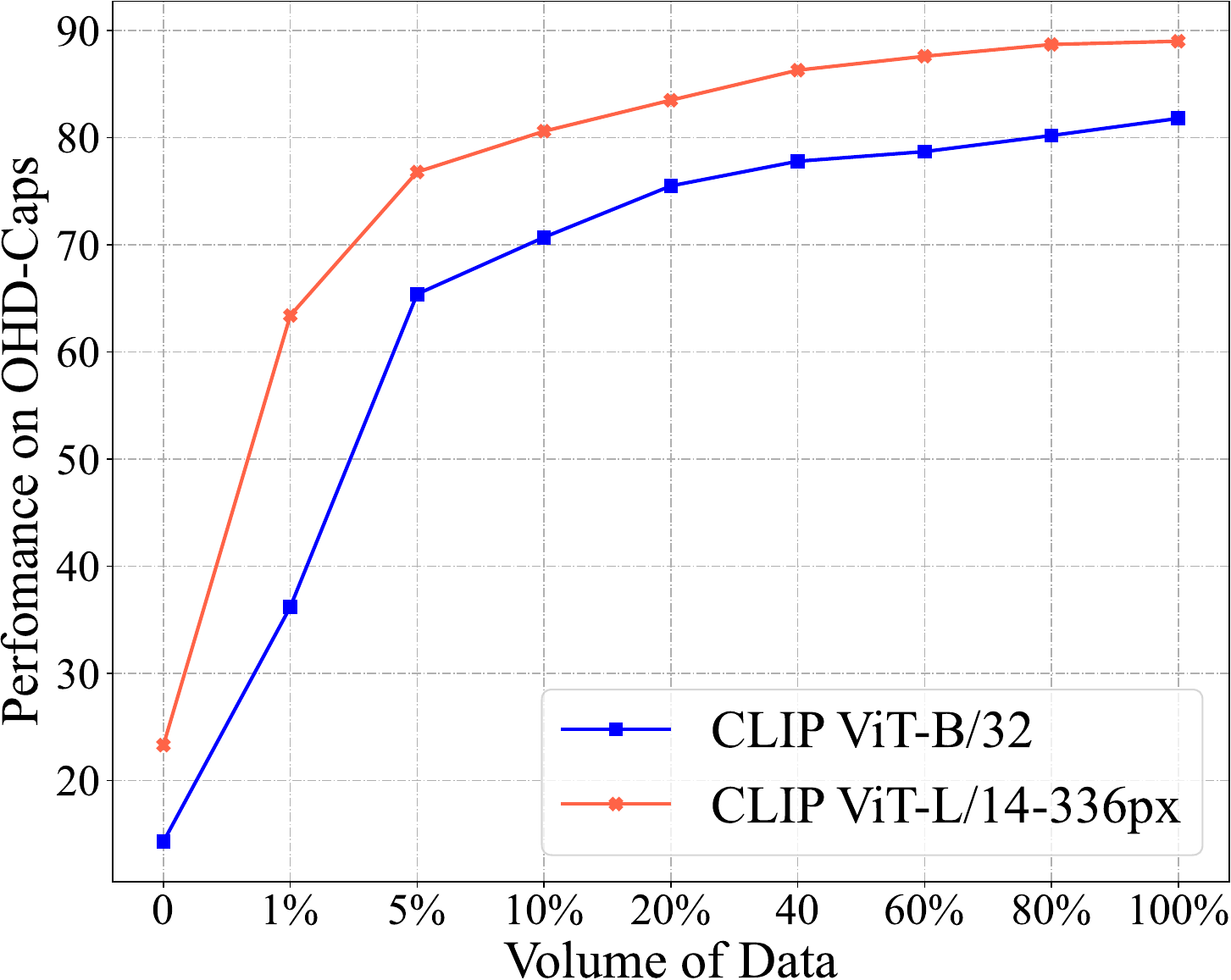}
    \caption{The performance of the model on the OHD-Caps dataset with different training data volumes provided. We report the average results of three random seeds.}
    \label{fig:data_num}
\end{figure}

In this subsection, we present ablation studies to examine the impact of our model's different components.
We conduct these experiments on the CLIP ViT-B/32 model.

\paragraph{Losses}
As demonstrated in Table \ref{tab:ablation_loss}, the inclusion of the $\mathcal{L}_0$ loss alone significantly improves OHD-Caps performance over the baseline. Subsequently, iterative incorporation of $\mathcal{L}_1$ and $\mathcal{L}_2$ provide incremental benefits, with the full combination yielding the highest average performance. 
Compared to $\mathcal{L}_1$ loss, $\mathcal{L}_2$ loss has a more significant effect on improving model performance. 
This suggests that by increasing the distance between constructed negative samples and other negative samples in the batch, the model can achieve a more refined understanding.

\paragraph{Data Volume}
Figure \ref{fig:data_num} shows the performance of the OHD-Caps dataset with varying amounts of training data. As can be seen from the figure, even with a very small amount of data, the model's performance can be significantly improved. For example, by training with just 1\% of the data (that is, 160 images), the performance of the CLIP-L/14 model can increase from 20\% to 60\%. However, as more data is added, the performance improvement gradually slows and stabilizes.


\section{Conclusion}
Our study investigates the reasons behind object hallucination in LVLMs. 
We construct a benchmark specifically for the evaluation of hallucinations and find that the visual perception module commonly used in current LVLMS, i.e., the CLIP model, cannot effectively discriminate hallucinated text. 
By designing negative samples and optimizing the contrastive loss function, we achieve a significant improvement in model performance on the hallucination detection dataset. 
Moreover, replacing the original CLIP model with our improved model can effectively alleviate the issue of object hallucination in the LLaVA model.

\section*{Limitations}
Although we conducted a series of explorations, our research still has its limitations. 
Firstly, our focus is solely on the issue of object hallucination within LVLMs, and we do not extend our research to other types of hallucinations. 
Secondly, the benchmark we propose comprises over 20 negative samples. 
Due to budgetary constraints, the size of this dataset is much smaller compared to the datasets used for evaluating compositional understanding, e.g. ARO dataset~\cite{DBLP:conf/iclr/Yuksekgonul0KJ023}.
Thirdly, we only evaluate the visual encoders of most LVLMs, i.e. the CLIP models, but we do not conduct research on encoders used by some other models, for instance, the variant of ResNet called NFNet-F6~\cite{DBLP:conf/icml/BrockDSS21} used by Flamingo~\cite{DBLP:conf/nips/AlayracDLMBHLMM22}.

\section*{Ethics Statement}
Object hallucination severely limits the practical application of LVLMs. For example, in medical image diagnosis, it can lead to false descriptions of tumor objects that are not present in the image. While our work has mitigated hallucinations in the visual encoder of LVLMs, hallucinations may still exist in the multi-head attention layers and feed-forward layers. Real-world applications based on LVLMs must systematically control hallucinations to avoid negative impacts on users.

\section*{Acknowledgement}
The authors wish to thank all reviewers for their
helpful comments and suggestions. The corresponding authors are Yuanbin Wu and Aimin Zhou. This research was (partially) supported
by NSFC(62076097), National Key R\&D Program
of China (2021YFC3340700), the Open Research
Fund of Key Laboratory of Advanced Theory and
Application in Statistics and Data Science (East
China Normal University), Ministry of Education.

\bibliography{custom}

\appendix
\section{Statistics on the Datasets}

\begin{table}[!htbp]
    \centering
    \resizebox{\columnwidth}{!}{
    \begin{tabular}{lccc}
        \toprule
        \bf Dataset & \bf Size & \bf \#Negative Samples& \bf \#Avg Length \\
        \midrule
        \rowcolor{gray!10} \multicolumn{4}{c}{\textit{\textbf{Train}}} \\
        COCO & 8000 & 27 & 16.0\\
        Flickr30K & 8000 & 27 & 18.4\\
        \rowcolor{gray!10} \multicolumn{4}{c}{\textit{\textbf{Dev}}} \\
        COCO & 990 & 27 & 15.6\\
        \rowcolor{gray!10} \multicolumn{4}{c}{\textit{\textbf{Test}}} \\
        COCO & 500 & 27 & 16.3\\
        Flickr30K & 500 & 27 & 21.1\\
        Nocaps & 500 & 27 & 19.1\\
        \bottomrule
    \end{tabular}
    }
    \caption{Statistics of the datasets used in our benchmark.}
    \label{tab:dataset}
\end{table}

The statistical information of the dataset is presented in the Table ~\ref{tab:dataset}, which is divided into three parts: training, testing, and validation. The average length displayed in the table refers to the average length of the negative examples in the dataset.

\begin{table}[t]
    \centering
    \setlength{\tabcolsep}{3pt}
    \resizebox{\linewidth}{!}{  
    \begin{tabular}{lcccccc}
    \toprule
    \multirow{2}{*}{\bf Model} &\multicolumn{2}{c}{\bf COCO }&\multicolumn{2}{c}{\bf Flickr30K }&\multicolumn{2}{c}{\bf Nocaps } \\
    \cmidrule(lr){2-3} \cmidrule(lr){4-5}\cmidrule(lr){6-7}
    & \bf F1 &\bf Yes \%  &\bf F1 &\bf Yes \% & \bf F1 &\bf Yes \% \\
    \midrule
     \multicolumn{7}{c}{(a) Full FT} \\
     \hline
    LLaVA   &86.4&56.5&79.2&73.1&80.2&66.0  \\
    LVIS-619k &77.4&32.6&70.2&33.6&67.3&31.2  \\
    LVIS-880k &85.6&41.7&79.7&\bf45.6&80.6&43.7  \\	
    \textbf{Ours} &\bf87.9&\bf46.9&\bf82.7&56.8&\bf82.0&\bf52.7 \\
    \hline
    \multicolumn{7}{c}{(a) LoRA FT} \\
    \hline
    LLaVA   &86.7&56.8&79.5&72.5&80.2&65.6  \\
    POVID &86.8&44.9&81.9&\bf51.8&81.4&49.6  \\	
    HADPO &84.6&43.0&75.1&43.5&78.4&43.7  \\	
    \textbf{Ours} &\bf88.2&\bf48.6&\bf83.5&52.9&\bf83.2&\bf52.3 \\
    \bottomrule
    \end{tabular}
    }
    \caption{Comparison results on expanded POPE datasets. Yes$\%$ denotes the proportion of answering ``Yes" to the given question.}
    \label{tab:compare_mthods_pope}
\end{table}

\section{Comparison with Other Methods} \label{sec:compare_sec}
To demonstrate that the proposed method has fewer object hallucinations and better general performance than other popular methods, we additionally compared the following approaches: LVIS~\cite{DBLP:journals/corr/abs-2311-07574} built a 220k visual instruction dataset. By utilizing the excellent visual analysis ability of GPT-4V and generating data through carefully designed prompts. Expanding the original LLaVA training data, datasets of different sizes, 619k and 880k, were obtained; POVID~\cite{DBLP:journals/corr/abs-2402-11411} and DPO~\cite{DBLP:journals/corr/abs-2311-16839} build hallucination texts using GPT4V and GPT4 respectively, and compose pairs with high-quality non-illusionary replies for DPO optimization. We report the model results based on the checkpoints provided by the paper.

\begin{table}[htbp] 
    \centering
    \setlength{\tabcolsep}{3pt}
    \resizebox{\columnwidth}{!}{ 
    \begin{tabular}{lccccc}
    \toprule 
    \bf Model & \bf MME & \bf VQAv2 & \bf VisWiz & \bf SciQA-IMG & \bf TextVQA \\
    \midrule
    \multicolumn{6}{c}{(a) Full FT} \\
    \hline
     LLaVA & 1459.4&79.1&48.9&\bf69.4&58.5 \\
     LVIS-619k &1473.6&79.2&50.0&68.1&57.7\\
     LVIS-880k &1517.7&\bf79.6&\bf51.7&68.9&\bf58.7 \\
    \bf Ours  &\bf1487.2&79.2&50.0&69.3&58.2 \\
    \midrule
    \multicolumn{6}{c}{(b) LoRA FT} \\
    \hline
    LLaVA &1445.4&79.1&46.8&69.8&\bf58.5  \\
    POVID&1418.5&78.8&42.3&67.5&58.0\\
     HADPO&1430.4&76.4&43.4&\bf70.3&56.6\\
    \bf Ours  &\bf1455.4&\bf79.2&\bf47.2&68&58.4 \\
    \bottomrule
    \end{tabular}
    }
    \caption{Comparison Results on various benchmarks.}
    \label{tab:compare_other_bnechmark}
\end{table}

The results are shown in Table \ref{tab:compare_mthods_pope} and Table \ref{tab:compare_other_bnechmark}.
From the results, our method outperforms the instruction finetune-based and dpo-based methods in terms of performance on POPE (our method improved the average F1 score by 2.6, while LVIS, HADPO, and POVID showed no significant improvement), demonstrating lower hallucination rates. Additionally, our method shows comparable performance to other methods in terms of general performance.

\section{More Examples}

\begin{figure*}[t]
    \centering
    \includegraphics[width=0.9\textwidth]{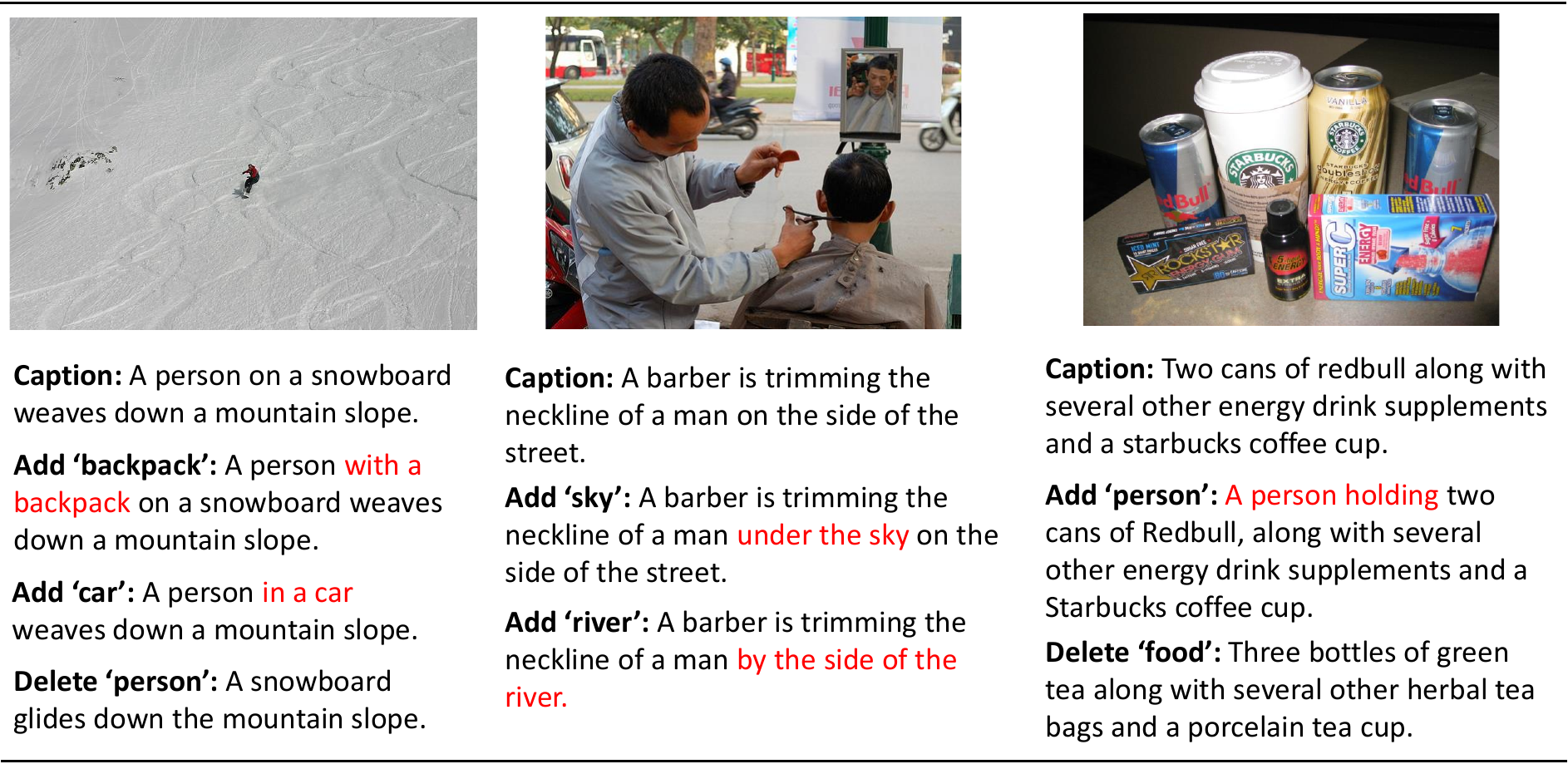}
    \caption{Examples from our benchmark OHD-Caps. The three images in the figure are from the COCO, Flickr, and Nocaps datasets, respectively.}
    \label{fig:exampels}
\end{figure*}

We present more examples in Figure~\ref{fig:exampels}.
It can be observed that our method can seamlessly integrate objects that are not present in the original image into the text. 
The names of the added objects are highlighted in red. 
Removing objects that are present in the picture can be accomplished with minimal adjustments. 
As for the removal of objects not depicted in the image, such as the ``food'' mentioned in the third figure, the negative samples typically involve modifications to the objects, attributes, and other content in the positive samples.

\begin{table*}[t]
    \centering
    \small
    \resizebox{0.9\linewidth}{!}{
    \begin{tabular}{m{11cm}}
    \toprule
        \bf Prompt Template \\
        \midrule
            \textbf{Add\_Prompt}: \textit{Given a sentence \{caption\}, generate a new sentence and includes each object from the list \{objects\}. Make the changes to the original sentence as minimal as possible. Ensure that the new sentence is coherent, natural, semantically smooth and free of grammatical errors.}\\
         \midrule
          \textbf{Remove\_Object\_Prompt}: \textit{Given a sentence \{caption\}, generate a new sentence and remove each object from list \{objects\} to make the semantics of the sentence different. Ensure that the new sentence is coherent, natural, semantically smooth and free of grammatical errors.} \\
         \midrule
          \textbf{Alter\_Object\_Prompt}: \textit{Given a sentence \{caption\}, choose to modify the objects, colors, attributes, etc., within the sentence to make the semantics of the sentence different. Make the changes to the original sentence as minimal as possible. Ensure that the new sentence is coherent, natural, semantically smooth and free of grammatical errors.}\\
    \bottomrule
    \end{tabular}
    }
    \caption{Prompt Templates for Querying GPT-4. We replace the object that is to be added or deleted with {object} in the prompt, and replace {caption} with the original caption text. The revised text should then be submitted to GPT-4 to generate the corresponding output.}
    \label{tab:gpt4_prompt}
\end{table*}

\section{Prompt Template}

Table~\ref{tab:gpt4_prompt} presents the prompt templates for generating negative samples that we used in Section~\ref{section:benchmarks}.



\end{document}